\documentclass{article}

\usepackage{PRIMEarxiv}

\usepackage[utf8]{inputenc} 
\usepackage[T1]{fontenc}    
\usepackage{amsmath}
\usepackage{booktabs}       
\usepackage{microtype}      
\usepackage{graphicx}
\usepackage{fancyhdr}       
\usepackage[numbers,sort&compress]{natbib}
\usepackage{url}
\usepackage{hyperref}       
\hypersetup{colorlinks=true, linkcolor=blue, citecolor=blue, urlcolor=blue}

\newcommand{\doi}[1]{\href{https://doi.org/#1}{doi:#1}}
\graphicspath{{images/}}

\title{Beyond Accuracy: How Procedural Traces Shift\\
the Decision Criterion of LLM Overseers
\thanks{Preprint. Accepted at the 60th Hawaii International Conference on System Sciences (HICSS), Hawaii, US.}}

\author{
  Zihan Chen \\
  Stevens Institute of Technology \\
  Hoboken, NJ, USA \\
  \texttt{zchen61@stevens.edu} \\
  \And
  Di Zhu \\
  Stevens Institute of Technology \\
  Hoboken, NJ, USA \\
  \texttt{dzhu1@stevens.edu} \\
  \AND
  Lei Zheng \\
  University of Massachusetts Boston \\
  Boston, MA, USA \\
  \texttt{lei.zheng@umb.edu} \\
  \And
  Weiling Li \\
  Stony Brook University \\
  Stony Brook, NY, USA \\
  \texttt{weiling.li@stonybrook.edu} \\
}

\begin{document}
\maketitle
\setcounter{footnote}{0}

\begin{abstract}
Organizations increasingly use oversight loops where one large language model (LLM) audits another's outputs alongside procedural traces of claimed steps. A common concern about such LLM-as-a-judge pipelines is that detailed traces make overseers gullible. Using signal detection theory, we audit five LLM overseers on 19 compliance tasks (4{,}551 analyzed judgments), varying only trace detail and evidence labeling. With disconfirming evidence always visible, error detection remains near ceiling. Instead, elaborate traces shift the decision criterion toward rejection, increasing false alarms in susceptible overseers. Without option labels, human-validated reason coding shows about 60\% of false alarms cite an inability to tie evidence to its option. Labels eliminate this stated reason, yet residual rejection of correct work persists in those overseers and rises with trace detail. Procedural traces thus act as governance artifacts that shape oversight decisions. AI auditors should be evaluated by their decision criterion and false-alarm behavior, alongside accuracy.
\end{abstract}


\section{Introduction}
``Let AI check AI'' has become a common safeguard. As generative systems move into high-stakes knowledge work, organizations increasingly experiment with oversight loops in which a second LLM judges, verifies, or audits the first. These pipelines now handle code review, content moderation, retrieval-augmented question answering, and compliance review, and ``verifier'' and ``critic'' personas are widely adopted in multi-agent workflows \cite{zheng2023judging}. The premise is that an automated overseer scales human review. Organizations adopt this arrangement out of necessity, because generative systems produce work far faster than people can check it \cite{berente2021managing}. GitHub Copilot users, for instance, complete tasks about 56\% faster \cite{peng2023impact}. In organizational terms, this loop is an automated control system. The trace attached to a worker model's output is more than an explanation: it is an input to a later control decision. Yet an overseer is only as good as its judgment, and that judgment depends on something little studied: the explanation bundled with the answer. How does an LLM overseer behave when the answer arrives with an account of how it was reached?

Modern agentic systems return an answer together with a \emph{procedural trace}. We use this term for a natural-language account, presented to the overseer, that claims the worker model decomposed the task, retrieved and cross-checked sources, and verified its recommendation \cite{singh2025agentic}. A concise trace for a restaurant recommendation might state ``I reviewed the criteria, checked the candidates against the reviews, and selected the best match,'' whereas a detailed trace walks through each requirement in turn. Such a trace is a presentational device, not necessarily the model's hidden chain of thought or a faithful execution log, and when one LLM oversees another it becomes the primary input the overseer acts on. A trace is a narrative of due diligence rather than evidence. It tells the overseer that the checking has already been done. Because such accounts need not reflect what a model actually did \cite{turpin2023language}, a trace can appear rigorous whether or not the underlying work was solid.

How should a longer, more ``rigorous'' trace affect an LLM overseer? It could cut two ways. It may induce overtrust: a confident ``I verified every criterion'' rationale could lead the overseer to treat the audit as complete and wave flawed outputs through, the gullibility that troubles practitioners and echoes automation bias in human-AI collaboration research \cite{skitka1999automation, parasuraman2010complacency, bansal2021whole}. Or a fuller trace may raise the overseer's guard: more asserted steps mean more claims to check, so a suspicious auditor may demand more before approving and reject more work, including work that is in fact correct. Both directions are plausible, which makes this an empirical question.

Existing work cannot adjudicate between these possibilities. The LLM-as-a-judge literature scores an overseer by accuracy or agreement with a reference and documents biases such as position, verbosity, and sycophancy \cite{zheng2023judging, wang2023fair, sharma2024sycophancy}. A single accuracy score, however, conflates two quantities long separated in signal detection theory \cite{green1966signal, macmillan2005detection}: sensitivity (can the overseer tell correct work from flawed work?) and response bias (how much doubt does it require before it rejects?). A judge that rejects more is not necessarily better; it may simply be more suspicious. The literature on explainable AI and automation bias draws exactly this distinction \cite{lee2004trust, vasconcelos2023explanations}, and we review that human-oversight evidence next. To our knowledge it has not been used to isolate how procedural traces affect an LLM overseer when answer quality and evidence are held fixed. That is our setting.

In this study, we ask: \textbf{(RQ1)} Does procedural-trace elaboration reduce an LLM overseer's error-detection sensitivity? \textbf{(RQ2)} Does it shift the overseer's decision criterion toward rejection, inflating false rejection of correct work? \textbf{(RQ3)} Is any such shift driven by the overseer's inability to tell which option a given piece of evidence describes, and does it persist once that link is supplied? We call this link \emph{evidence-to-option attribution}, or attribution for short. It is a property of how evidence is displayed to the overseer, not of the trace, and it is narrower than provenance in the usual sense of document lineage. \textbf{(RQ4)} How do these effects vary across overseer models?

We answer these questions with a controlled, fully crossed audit grounded in signal detection theory (Section~\ref{sec:method}). Five LLM overseers each judge 19 compliance-style tasks in which an assistant recommends an option that, by construction, is either correct or contains a planted error. Holding the recommendation and the underlying evidence fixed, we vary only the trace (four levels of increasing ``performed rigor,'' up to a ``flawed-verification'' trace that falsely claims to have checked the violated requirement) and whether each review is labeled with the option it describes. Crossing five models, 19 items, two correctness levels, four traces, two attribution conditions, and three runs yields 4{,}560 judgments, of which 4{,}551 parsed cleanly and are analyzed. Because only the presentation changes within a task, behavior changes are attributable to the trace and the labeling rather than to answer quality, and the item is the unit we cluster on throughout.

The headline result runs opposite to the gullibility worry. Detection never degrades: pooled hit rates run from 99.3\% at no trace to 100\% at the most detailed one, and 98\% of those catches name the criterion actually violated, so the ceiling reflects genuine detection rather than indiscriminate rejection. Because errors are single-criterion violations whose disconfirming review is always on screen, this establishes that a trace does not override available evidence; it does not speak to concealed errors. Yet the decision criterion slides toward rejection (the criterion $c$ and sensitivity $d'$ are defined in Section~\ref{sec:theory}). In the most susceptible overseer, wrongly rejecting correct answers climbs from 58\% to 96\% as the trace grows more detailed, and pooled across overseers each step up the trace ladder raises the odds of wrongly rejecting correct work by about 44\%. Labeling each review with the option it describes cuts this sharply but does not remove it, and the surviving part generalizes furthest: a rigor-driven over-skepticism persists even under best-practice attribution. The effect is also heterogeneous, concentrated in the three open-weight models while the two frontier models barely move, which is why we treat overseer calibration as a moderator rather than assuming a universal effect.

The practical risk, then, is that explanations can create unwarranted suspicion as well as unwarranted trust. The broader implication for information systems is that procedural traces, which double as the audit trail attached to an AI system's output, are governance artifacts rather than passive transparency devices. They can move the operating point of a downstream control, the balance it strikes between missing errors and false alarms. Our contributions follow. \emph{Theoretically and methodologically}, we reframe LLM oversight through signal detection theory, separating an overseer's sensitivity from its decision criterion in a way accuracy-only evaluations cannot; we introduce a controlled paradigm that treats the LLM overseer as the experimental subject and holds the audited answer and evidence fixed; and we decompose ``performed rigor'' into a removable attribution gap and a residual, overseer-specific over-skepticism. \emph{For practice and design}, procedural traces can recalibrate a downstream overseer even when the answer and evidence are unchanged, labeling evidence with the option it describes is a cheap partial remedy, and AI auditors should be governed by their decision criterion and false-alarm behavior rather than by accuracy alone.

\section{Related work}

\textbf{LLM-as-a-judge and AI oversight.} Using LLMs to evaluate or supervise other models has become standard, from benchmark scoring to iterative self-critique \cite{zheng2023judging, chen2026survey}. This line documents a growing catalogue of systematic judge biases: position and verbosity effects \cite{wang2023fair}, self-preference, the tendency to rate one's own outputs higher because they are more familiar \cite{wataoka2024self}, and sycophancy, the tendency to agree with a stated view \cite{sharma2024sycophancy}. The failure we report joins that catalogue, but it is not a preference artifact: it moves where the judge sets its threshold. These studies mostly examine preference or quality judgments and summarize the judge by a single accuracy metric. We depart in two ways. We study binary oversight decisions over correct versus erroneous recommendations, and we model the judge as a detector with two separable parameters rather than one accuracy number, using that lens to isolate how procedural traces move the criterion while answer quality is held fixed.

\textbf{Explanations, overreliance, and automation bias.} A large literature on human-AI decision-making shows that explanations are a double-edged sword. They can improve appropriate reliance, but they can also inflate confidence and induce overreliance without improving decisions \cite{bansal2021whole, vasconcelos2023explanations}, building on classic results on automation bias, complacency, and calibrated trust \cite{skitka1999automation, parasuraman2010complacency, lee2004trust}. It is also where the sensitivity-versus-criterion distinction originates: decades of detection and monitoring research separate an operator's \emph{sensitivity} from \emph{response bias}, and find that decision aids can shift the bias without changing sensitivity \cite{green1966signal, macmillan2005detection, parasuraman2010complacency}. That work motivates our central tension, but it concerns human overseers. Whether an AI overseer inherits the same susceptibility, and in which direction, remains an open question.

\textbf{Retrieval, agentic traces, and faithfulness.} Agentic retrieval-augmented generation surfaces planning, tool use, and self-verification as procedural traces \cite{singh2025agentic}, and such traces need not faithfully reflect a model's actual computation \cite{turpin2023language}. Prior faithfulness work asks whether a trace truthfully reports the generator's own reasoning; we ask how its presence alters another model's decision threshold.

\textbf{Procedural traces as governance artifacts.} For information systems, the more useful framing is organizational. Procedural traces function as governance artifacts: the audit trails on which a later control decision is built. They do not simply reveal reasoning; they redistribute perceived verification responsibility between the worker model and the overseer, so the question becomes whether such a trace recalibrates the downstream evaluator. This connects to the IS control literature, which distinguishes the controls an organization configures from how they are enacted \cite{wiener2016control}, and to work showing control increasingly delegated to algorithms \cite{kellogg2020algorithms, mohlmann2021algorithmic, gao2024open}. It also inherits a known difficulty: evaluating AI-produced work is itself unreliable, because the ground truth and accountability records the overseer leans on are often weaker than they appear \cite{lebovitz2021ground, asatiani2021sociotechnical}.

\section{Theory and hypotheses}\label{sec:theory}

We model the overseer as a binary detector that decides whether to ACCEPT or REJECT an audited recommendation. Signal detection theory \cite{green1966signal, macmillan2005detection} separates two quantities. \emph{Sensitivity} ($d'$) measures how well the overseer distinguishes a flawed recommendation from a sound one, and it rises with the gap between hits (correctly rejecting an error) and false alarms (rejecting a correct answer). The \emph{criterion} ($c$) measures response bias, how much evidence of a problem the overseer requires before it rejects, where more negative $c$ denotes a stronger predisposition to reject. Put plainly, $d'$ is how far apart flawed and sound work look to the overseer, and $c$ is where the overseer draws the line between them; Figure~\ref{fig:sdt} draws both from our own estimates. A judge's accuracy can therefore move because it sees the difference better or because it has become more suspicious, two cases with opposite design implications that a single accuracy score cannot distinguish.

The competing forces map onto these parameters as two channels. The gullibility channel predicts that performed rigor lowers \emph{sensitivity}, so a convincing trace makes errors harder to catch. \textbf{H1a (gullibility):} more elaborate traces reduce $d'$ and lower the hit rate on planted errors. The over-skepticism channel predicts that performed rigor instead shifts the \emph{criterion}, since more asserted steps invite more doubt. \textbf{H1b (criterion shift):} longer traces shift the criterion toward rejection (more negative $c$) and raise the false-alarm rate on correct work, without necessarily reducing $d'$. The two channels are not mutually exclusive; our design measures which one dominates. In the language of IS control, the trace leaves the control an organization \emph{configures} untouched, namely the accept/reject task it assigns, while moving the control the overseer \emph{enacts}, namely where it sets its rejection threshold \cite{wiener2016control}. Performed rigor thus acts on control enactment, not configuration.

Why would a longer trace breed doubt? A detailed trace asserts more verifiable claims, and an overseer that tries to confirm them can fail in two ways. First, it may be unable to tell which option a piece of evidence describes, an attribution gap in how evidence is displayed rather than a defect in the recommendation. \textbf{H2 (attribution):} a substantial part of the criterion shift is this verifiability gap, so labeling each excerpt with the option it describes reduces false rejection of correct work. Second, even with that label supplied, an overseer may treat ``not affirmatively confirmed'' as ``wrong.'' \textbf{H3 (residual rigor):} a criterion shift driven by performed rigor persists even when evidence is fully attributed. Finally, calibration is a property of the individual overseer. \textbf{H4 (heterogeneity):} the criterion shift varies across overseers, with some robust in both attribution conditions.

The model is therefore one cause with two paths. Added trace detail raises the overseer's verification burden, which shifts the criterion toward rejection through an attribution gap and an absence-of-evidence rule, inflating false alarms on correct work, with attribution, overseer calibration, and evidence visibility as moderators.

\section{Method}\label{sec:method}
We operationalize the competing-forces test as a controlled audit: an LLM overseer judges recommendations whose correctness is fixed by construction, while we vary only the displayed trace and the way evidence is labeled.

\subsection{Task and stimuli}
Each task is a compliance-style recommendation audit. A user states a request with several explicit criteria that must all be satisfied (e.g., choose an option that is budget-friendly, with good food and good service); the overseer sees the candidate options, a pool of short evidence excerpts (customer reviews) drawn from the Yelp Open Dataset\footnote{\url{https://www.yelp.com/dataset}. Accessed 2026.}, and an assistant's recommendation. Real Yelp reviews of restaurants in three U.S. cities are assembled into a hand-validated set of 19 compliance tasks, the only data input to the audit. Each item has a known ground truth: a \emph{best} option satisfying all criteria and an \emph{error} option violating exactly one criterion on an objective, structured attribute. Each carries two recommendation variants naming the best and the violating option, and the evidence pool always contains the excerpt that reveals the violation. Because that disconfirming evidence is always visible, a careful overseer can detect every planted error, so failures of detection cannot be blamed on missing information. Figure~\ref{fig:stim} shows one item in full.

\subsection{Manipulations}
We manipulate two factors within each item, holding the recommendation and the evidence pool fixed. \textbf{Trace} has four ordered levels of increasing ``performed rigor.'' \emph{None} shows the recommendation alone. \emph{Concise} adds a four-step summary. \emph{Detailed} gives a per-criterion walkthrough with explicit self-verification language. \emph{Flawed-verification} adds to it an explicit, and for error items false, claim of having verified the very criterion that is violated. At every level the trace argues for the shown recommendation, and it is generated from the item by template, so it is identical across all five overseers and tailored to none of them. The levels grow in length as rigor increases, and Section~5.7 shows the effect tracks asserted verification rather than length alone. \textbf{Attribution} has two levels. \emph{Unattributed} shows each evidence excerpt with its feature and sentiment but not which option it describes. \emph{Attributed} additionally tags each excerpt with the option it is about, changing only the label on the evidence and never its content. The unattributed condition models a common failure mode in organizational AI pipelines, where evidence is summarized or aggregated by criterion in ways that preserve content but strip the link to the entity it describes, a known weakness of automated source attribution \cite{gao2023enabling}. A compliance dashboard that rolls every ``price'' complaint into one panel keeps the text but loses which restaurant each concerned, and so does any retrieval layer that concatenates snippets without carrying their subject forward. The manipulation therefore tests whether binding evidence to an option, not merely making it available, is needed for calibrated oversight. Within an item, candidate and evidence order are identical across conditions and cannot confound the comparison.

\begin{figure}[tb]
\centering
\fbox{\begin{minipage}{0.96\linewidth}
\footnotesize\raggedright\setlength{\parindent}{0pt}
\textbf{One of the 19 items, verbatim.} Blocks identical in every condition:

\smallskip
\textbf{Request.} Best French restaurant in Nashville; the criteria \emph{budget}, \emph{food} and \emph{service} must all hold.\\
\textbf{Candidates.} Le Sel (price tier 2/4) $\cdot$ Chateau West (3/4) $\cdot$ Table 3 (2/4) $\cdot$ Cafe Fundamental (2/4)\\
\textbf{Evidence.} Seven review excerpts, among them the one that disconfirms \emph{budget}: ``\ldots laugh off that \$80 substandard meal for two\ldots''\\
\textbf{Recommendation.} \emph{Error} version: ``I recommend Chateau West'' (price tier 3/4 violates \emph{budget}). \emph{Correct} version: ``I recommend Le Sel.''

\smallskip\hrule\smallskip
\textbf{(a) Attribution}, two levels. One tag is the only difference:\\
\texttt{unattributed}: \texttt{[e1] (feature: price, sentiment: neg)} then the review text.\\
\texttt{attributed}: \texttt{[e1] (about: Chateau West, feature: price, sentiment: neg)} then the same review text.\\
Unattributed, none of the seven excerpts can be tied to any of the four candidates.

\smallskip
\textbf{(b) Trace}, four levels of asserted rigor:\\
\texttt{none}: the block is absent.\\
\texttt{concise}: four steps, ``1.~Parsed the user's criteria\ldots 4.~Selected Chateau West as the best overall match.''\\
\texttt{detailed}: seven steps, one per criterion, ``Step 3 - Checked `budget': reviewed the available evidence for Chateau West. Assessment: satisfies `budget'.''\\
\texttt{flawed-verification}: eight steps, adding one claim that is false on error items, ``Step 6 - Verification of `budget': I specifically re-examined the budget evidence and confirmed that Chateau West meets the user's `budget' requirement.''
\end{minipage}}
\caption{One audit item in full. Only the evidence tag (a) and the trace (b) vary; everything else is byte-identical across conditions.}
\label{fig:stim}
\end{figure}

\subsection{Overseers, design, and procedure}
The design crosses 5 models, 19 items, 2 correctness levels, 4 traces, 2 attribution conditions, and 3 runs, for $5\times19\times2\times4\times2\times3 = 4{,}560$ judgments. We use five overseers chosen to span developers and weight-access regimes, so that the effect is not a quirk of one family: two frontier closed-weight Anthropic models (\texttt{claude-opus-4-6} and \texttt{claude-opus-4-7}) and three open-weight models, MiniMax \texttt{minimax-m2.5}, Google \texttt{gemma-3-27b-it}, and Alibaba \texttt{qwen3-next-80b-a3b}. We shorten these to \texttt{opus-4-6}, \texttt{opus-4-7}, \texttt{minimax-m2.5}, \texttt{gemma-3-27b}, and \texttt{qwen3-next-80b}.
Each overseer is instructed that it is an independent quality-control reviewer, that the assistant is sometimes wrong, and that it should verify claims against the evidence before deciding. It returns a structured ACCEPT or REJECT decision, a confidence rating from 0 to 100, a stated reason, and the evidence it relied on. All models were queried in a single run to limit version drift. Decoding used a fixed moderate temperature of 0.7 for the four models that accept it. \texttt{opus-4-7} rejects a temperature parameter and was run at its provider default, which the provider does not publish; Section~5.7 verifies that this does not account for its behavior. Because the three runs per cell share identical prompts and nondeterministic decoding, we treat them as repeated measurements and use the item as the clustering unit for all confirmatory inference. Of 4{,}560 judgments, 4{,}551 (99.8\%) parsed cleanly and enter the analysis; all reported rates, regressions, and odds ratios are computed on these 4{,}551, with the nine unparsed returns (0.2\%) dropped listwise. The 19 constructed items, the four trace templates, the overseer and judge prompts, and the analysis scripts will be released with the paper, since the Yelp source is public but the constructed stimuli are the binding constraint on reproduction.

\subsection{Measures and identification}
For each (model $\times$ trace $\times$ attribution) cell we compute the \emph{hit rate} (errors correctly rejected), the \emph{false-alarm rate} (correct recommendations wrongly rejected), and the derived SDT quantities $d'$ and $c$, applying the standard log-linear correction for extreme rates \cite{hautus1995corrections}. Because a hit could in principle be an accident of a generally suspicious overseer, we also record whether a rejected error names the criterion that was actually violated, which we call a \emph{right-reason hit}. We use \emph{sensitivity} throughout in its signal-detection sense, $d'$, not the machine-learning sense of recall. On our REJECT-is-positive convention the two error rates map to the standard SDT cells: a 99\% hit rate on error items means a 1\% miss rate, and a 58\% false-alarm rate means 58\% of correct recommendations were wrongly rejected. Identification rests on the held-fixed design. Within an item the recommendation and the evidence pool are constant across conditions, so differences in ACCEPT/REJECT behavior isolate the effect of the display. For inference we fit logistic regressions of the false-rejection indicator on the ordinal trace ``dose,'' the attribution condition, and their interaction, with standard errors clustered on item.

To characterize why an overseer rejects a correct recommendation, we classify each of the 934 stated reasons into one of four categories: \emph{attribution gap} (the overseer cannot tell which option an excerpt describes), \emph{over-skepticism} (it rejects because a criterion is not affirmatively confirmed), \emph{substantive} (it cites genuine disconfirming evidence), and \emph{other}. Two LLM judges label all reasons independently, one a panel member and one (\texttt{claude-opus-4-8}) outside the panel, which guards against a model grading its own outputs. To check that the judges track people and not only each other, two human annotators label a stratified subsample of 103 reasons with the same rubric, blind to condition, model, and judge labels. We report inter-judge, judge-human, and inter-annotator agreement as Cohen's $\kappa$ (Section~5.5) and do not resolve disagreements, since $\kappa$ already quantifies them. A third early return was excluded before analysis as a near-duplicate ($\kappa=0.99$) and so not independent.

\section{Results}
We first show that error detection is unharmed (RQ1), then that the criterion shifts toward rejection (RQ2). We then decompose that shift into an attribution gap and a residual rigor effect (RQ3), and characterize the cross-model heterogeneity (RQ4).

\subsection{No gullibility: detection is intact (RQ1, H1a)}
Error detection remains at ceiling and does not fall as the trace grows. Pooled hit rates across the four trace levels are 99.3\%, 99.5\%, 100\%, and 100\%; the lowest model $\times$ trace cell is 97.4\%, and no cell of the full 40-cell design falls below 94.7\%. A longer, or even self-congratulatory, ``flawed-verification'' trace does not make overseers miss planted errors. The ceiling is genuine detection rather than a by-product of an overseer rejecting everything: 98.1\% of hits name the criterion actually violated, and that right-reason rate is itself flat across the trace ladder (98.4\%, 98.1\%, 97.5\%, 98.4\%). H1a is not supported here, and the result is precise about its regime. Errors are single-criterion violations and the disconfirming review is always on screen, so what we establish is that a trace does not make an overseer discard evidence it can see. Detection under concealed errors is a different question, taken up in Section~6.3.

\subsection{The criterion slides toward rejection (RQ2, H1b)}
\begin{figure}[tb]
  \centering
  \includegraphics[width=0.78\linewidth]{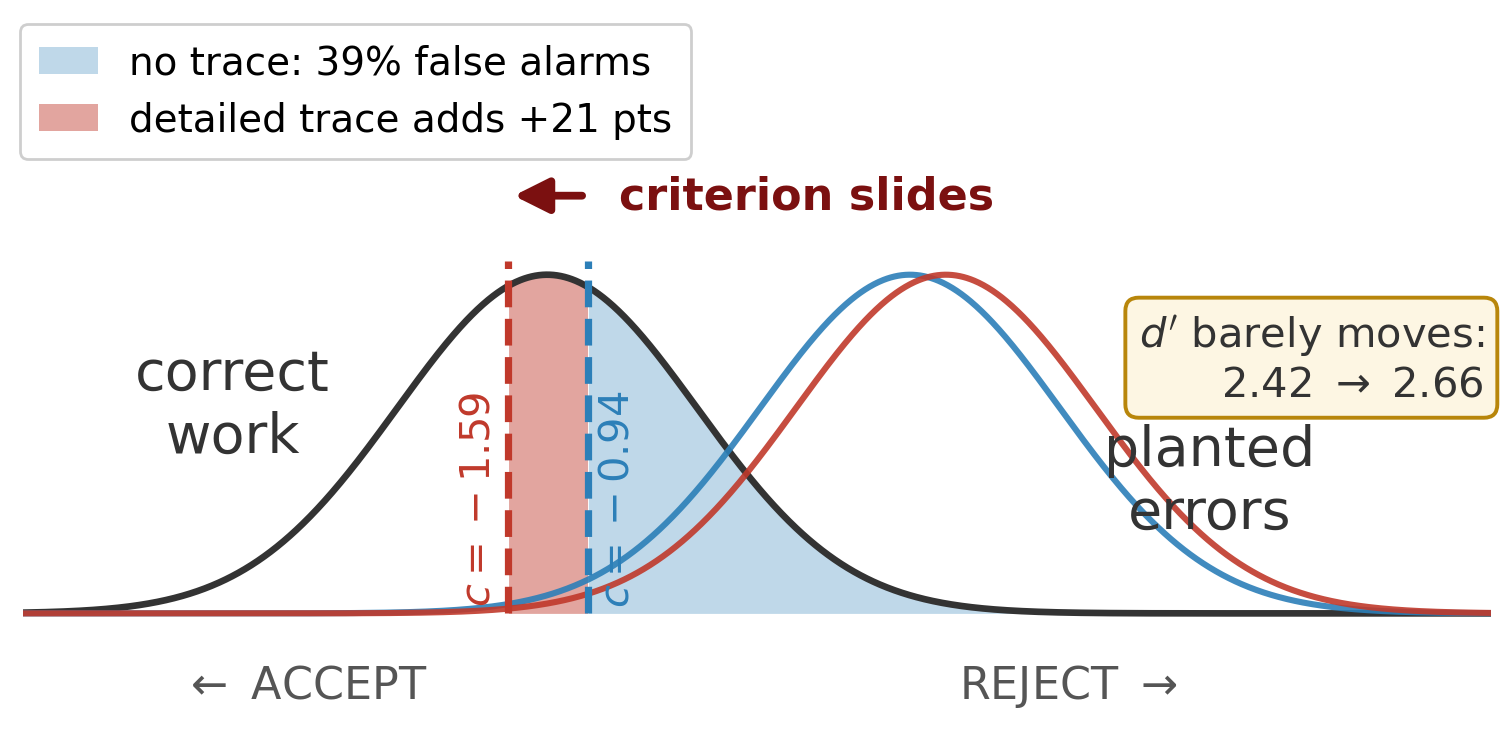}
  \caption{The criterion shift, drawn from the pooled estimates in Table~\ref{tab:acc} (unattributed arm). $d'$ is essentially unchanged, so the distributions barely move; the threshold slides toward REJECT and sweeps 21 more points of correct work into the false-alarm region.}
  \label{fig:sdt}
\end{figure}

\begin{figure}[!tb]
  \centering
  \includegraphics[width=0.98\linewidth]{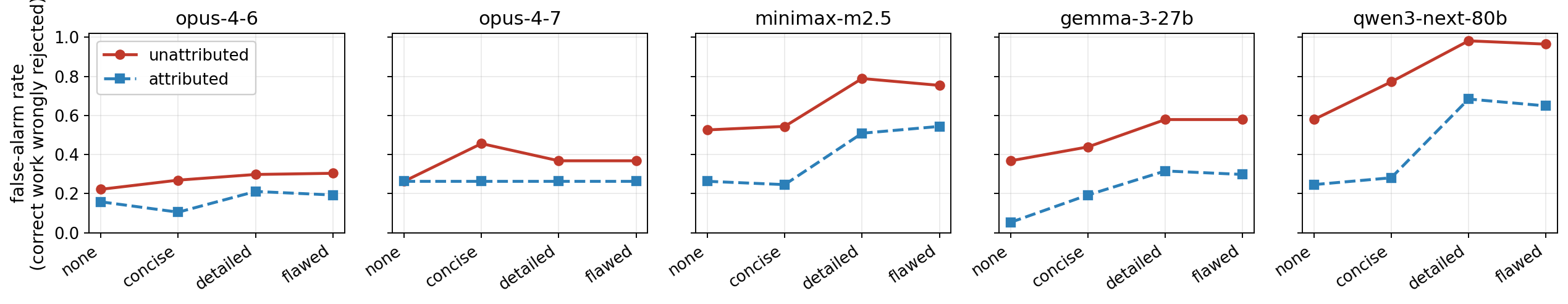}
  \caption{False-alarm rate by trace level (red: unattributed; blue: attributed). False alarms rise with trace detail in the susceptible models; attribution lowers the level but not the slope, and the \texttt{opus} controls are flat.}
  \label{fig:fa}
\end{figure}

While detection holds, the false-alarm rate on correct recommendations climbs steeply as the traces grow more detailed in the susceptible overseers (Figure~\ref{fig:fa}, solid lines). In the unattributed arm, moving from no trace to the flawed-verification trace raises false alarms from 58\% to 96\% for \texttt{qwen3-next-80b}, from 37\% to 58\% for \texttt{gemma-3-27b}, and from 53\% to 75\% for \texttt{minimax-m2.5}. The two \texttt{opus} overseers barely move (22\% to 30\% and 26\% to 37\%). In SDT terms, the criterion $c$ slides toward rejection exactly where false alarms rise, for example from $-1.29$ to $-2.05$ for \texttt{qwen3-next-80b} and from $-0.61$ to $-1.29$ for \texttt{gemma-3-27b}, while $d'$ does not collapse. Pooling all overseers (Table~\ref{tab:reg}, Panel A), each one-level increase in the trace dose multiplies the odds of wrongly rejecting a correct answer by 1.44 (95\% CI [1.28, 1.63]). This is the over-skepticism pattern of H1b, not gullibility: performed rigor moves the criterion, not competence. As the trace grows more detailed, pooled accuracy falls from 0.80 to 0.70 while $d'$ stays near 2.4 to 2.7, so the entire accuracy drop is the criterion sliding (Table~\ref{tab:acc}). Figure~\ref{fig:sdt} draws the same comparison in detection terms: the distributions stay where they are, the threshold slides left, and the false-alarm area under the correct-work distribution grows from 39\% to 60\%.

\subsection{Attribution removes a large component (RQ3, H2)}
Labeling each evidence excerpt with the option it describes sharply reduces over-rejection in every overseer (Figure~\ref{fig:fa}, dashed lines). At the flawed-verification trace, attribution cuts the false-alarm rate by about 32 points for \texttt{qwen3-next-80b} (96\% to 65\%), by 28 points for \texttt{gemma-3-27b} (58\% to 30\%), and by 21 points for \texttt{minimax-m2.5} (75\% to 54\%). Controlling for trace and model fixed effects, attribution multiplies the odds of false rejection by 0.34 (95\% CI [0.17, 0.67]; Table~\ref{tab:reg}, Panel A). H2 is supported: a substantial part of the criterion shift is a verifiability gap, the overseer rejecting because it cannot tell which option a review describes. That component is conditional on a specific and avoidable weakness, since a retrieval layer that carries each snippet's subject forward never creates it. This is why the intervention is cheap, and why the part that survives it, reported next, is the more general finding.

\begin{table}[tb]
\centering
\caption{Logistic regression of false rejection of correct work (DV $=1$ if rejected; cluster-robust SE on item). Panel~A pools with model fixed effects; Panel~B gives the trace-dose slope $\beta$ (dose none$=0$ to flawed$=3$).}
\label{tab:reg}
\small
\setlength{\tabcolsep}{8pt}
\begin{tabular}{@{}lcc@{}}
\toprule
\multicolumn{3}{@{}l}{\textbf{Panel A. Pooled effect}}\\
\midrule
Term & Odds ratio & 95\% CI \\
\midrule
Attribution (attributed) & 0.34 & [0.17, 0.67] \\
Trace dose (per level)   & 1.44 & [1.28, 1.63] \\
\addlinespace
\multicolumn{3}{@{}l}{\textbf{Panel B. Trace-dose slope $\beta$ [95\% CI]}}\\
\midrule
Model & Unattributed & Attributed \\
\midrule
\texttt{opus-4-6}       & $+0.14$ [$-0.18$, $0.46$] & $+0.15$ [$-0.08$, $0.38$] \\
\texttt{opus-4-7}       & $+0.10$ [$-0.21$, $0.41$] & $\phantom{+}0.00$ (flat) \\
\texttt{minimax-m2.5}   & $+0.42$ [$+0.16$, $0.69$] & $+0.48$ [$+0.21$, $0.76$] \\
\texttt{gemma-3-27b}    & $+0.31$ [$+0.02$, $0.61$] & $+0.54$ [$+0.19$, $0.89$] \\
\texttt{qwen3-next-80b} & $+1.22$ [$+0.54$, $1.90$] & $+0.70$ [$+0.32$, $1.08$] \\
\bottomrule
\end{tabular}
\end{table}

\subsection{A residual rigor effect survives attribution (RQ3, H3)}
Attribution lowers the \emph{level} of over-rejection but does not flatten its \emph{slope}. In the attributed arm, false alarms still climb as the trace grows more detailed (\texttt{gemma-3-27b} 5\% to 30\%, \texttt{minimax-m2.5} 26\% to 54\%, \texttt{qwen3-next-80b} 25\% to 65\%), and the trace-dose slope remains positive and significant for these three overseers ($p=.003$, $.001$, and $<.001$; Table~\ref{tab:reg}, Panel B), while the two \texttt{opus} overseers are flat in both arms. At the flawed-verification trace, the criterion $c$ moves back toward neutral under attribution but stays below baseline (\texttt{qwen3-next-80b} $-2.05$ to $-1.38$; \texttt{gemma-3-27b} $-1.29$ to $-0.93$). H3 is supported: a rigor-driven criterion shift exists independently of the attribution gap.

\begin{table}[!hbtp]
\centering
\caption{Accuracy hides the criterion shift: as trace detail increases (unattributed, pooled), accuracy falls while $d'$ stays flat and only $c$ moves.}
\label{tab:acc}
\small
\setlength{\tabcolsep}{8pt}
\begin{tabular}{@{}lccccc@{}}
\toprule
Trace & Accuracy & Hit & False alarm & $d'$ & $c$ \\
\midrule
none      & 0.80 & 0.99 & 0.39 & 2.42 & $-0.94$ \\
concise   & 0.75 & 0.99 & 0.50 & 2.25 & $-1.12$ \\
detailed  & 0.70 & 1.00 & 0.60 & 2.66 & $-1.59$ \\
flawed    & 0.70 & 1.00 & 0.60 & 2.68 & $-1.58$ \\
\bottomrule
\end{tabular}
\end{table}

\subsection{What the residual is, and heterogeneity (RQ4, H4)}
The overseers' stated reasons for wrongly rejecting correct work make the two parts concrete. Two independent LLM judges classified all 934 false-alarm reasons as attribution gap, over-skepticism, substantive, or other (Table~\ref{tab:mech}). They agree almost perfectly (Cohen's $\kappa=0.89$, $n=934$). Two human annotators on a stratified 103-item subsample agree with the judges at $\kappa$ between $0.71$ and $0.76$, and with each other at $\kappa=0.87$. The attribution-gap labels were human-confirmed 17 of 17 times for one judge and 18 of 19 for the other. The decomposition therefore does not rest on ``LLMs grading LLMs.''

\begin{table}[tb]
\centering
\caption{Stated-reason mix among false alarms (adjudicated; two judges A\,/\,B, with B off-panel; inter-judge $\kappa=0.89$).}
\label{tab:mech}
\small
\begin{tabular}{@{}lcc@{}}
\toprule
Stated reason & Unattributed & Attributed \\
              & (A\,/\,B)    & (A\,/\,B) \\
\midrule
Attribution gap  & 60\,/\,61\% & 0\,/\,0\% \\
Over-skepticism  & 11\,/\,10\% & 48\,/\,52\% \\
Substantive      & 29\,/\,29\% & 48\,/\,46\% \\
Other            & \phantom{0}1\,/\,0\%  & \phantom{0}4\,/\,2\% \\
\midrule
$n$ (false alarms) & 592 & 342 \\
\bottomrule
\end{tabular}
\end{table}

The attribution-gap category, meaning the overseer states it cannot tell which option an excerpt describes, is the single largest stated reason for over-rejection, about 60\% of false alarms in the unattributed arm, and it collapses to exactly zero under attribution for both judges. Attribution does not merely reduce the verifiability complaint; it removes it. What remains divides between genuine, if mistaken, substantive audits and an ``absence-of-evidence'' over-skepticism. There the overseer rejects correct work because a criterion is not affirmatively confirmed by any review, treating ``not proven'' as ``wrong.'' That over-skepticism is itself dose-responsive to the trace even when every excerpt is attributed: for \texttt{minimax-m2.5} and \texttt{qwen3-next-80b} the count of such rejections rises from no trace to the flawed-verification trace under both judges (11 to 22 and 3 to 15 for one, 10 to 19 and 2 to 14 for the other), while it is flat for the \texttt{opus} pair. The heterogeneity is stark. The \texttt{opus} pair is a robust control, flat in both arms, while \texttt{qwen3-next-80b} rejects nearly all correct work under a detailed unattributed trace. H4 is supported, as a difference among these five specific systems. They vary at once in scale, vendor, weight availability, and post-training, so what makes an overseer robust is a question for a larger panel (Section~6.3).

\subsection{Confidently wrong (RQ4)}
Over-rejection comes with no usable warning signal in the most severe overseer. When \texttt{qwen3-next-80b} wrongly rejects a correct recommendation ($n=294$) it reports mean confidence 94.4, against 94.9 when it correctly accepts. The gap is 0.5 points on a 0 to 100 scale (95\% CI [0.02, 0.95], item-clustered), detectable but far too small to act on. The contrast with the calibrated overseers is what matters: \texttt{gemma-3-27b} drops 15.7 points when it over-rejects (75.6 vs.\ 91.3; CI [12.8, 18.5]), \texttt{opus-4-6} drops 13.7 (66.1 vs.\ 79.8), and \texttt{minimax-m2.5} drops 6.6. In the overseers most prone to manufacturing false alarms, confidence is essentially the same whether the overseer is right or wrong, so a downstream gate could not threshold the failure away.

\subsection{Robustness}
\begin{figure}[tb]
  \centering
  \includegraphics[width=0.56\linewidth]{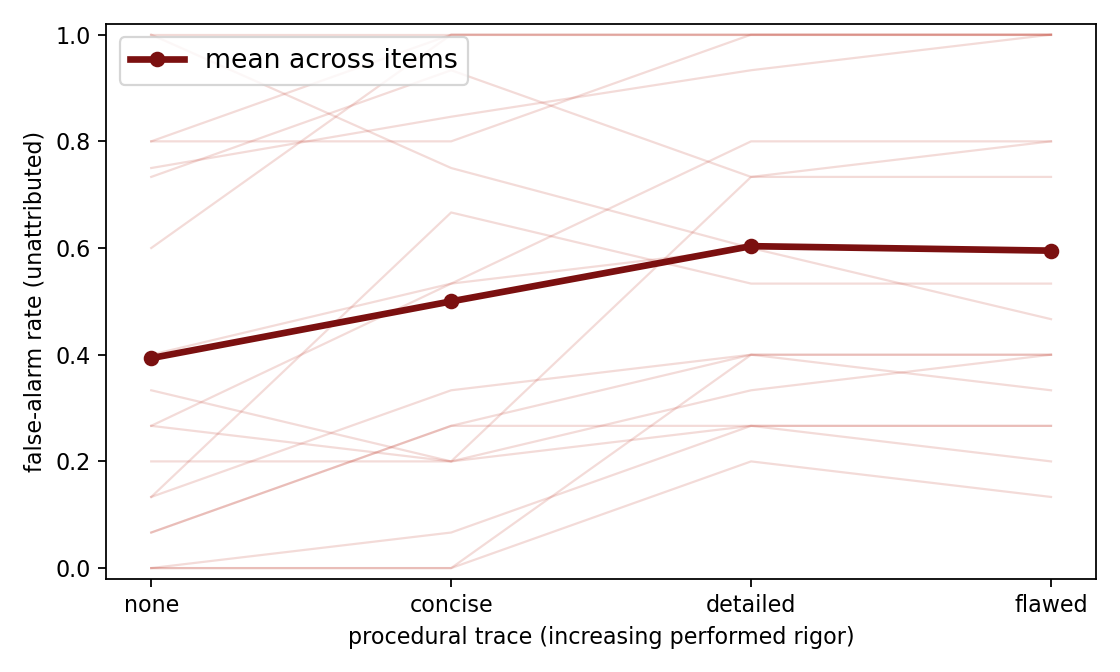}
  \caption{Per-item false-alarm rate by trace (unattributed; faint: the 19 items, bold: mean).}
  \label{fig:item}
\end{figure}

The findings hold under several alternative specifications. \emph{Categorical trace.} Treating trace as categorical rather than an ordinal dose gives the same picture: relative to no trace, the odds of false rejection rise to 1.40 at concise (95\% CI [1.07, 1.82]), 2.82 at detailed [1.95, 4.08], and 2.71 at flawed-verification [1.93, 3.81]. The rise concentrates from concise to detailed and then plateaus, which is inconsistent with an effect driven purely by length. \emph{Item robustness.} Leaving out one item at a time keeps both effects stable across all 19 specifications (trace-dose OR in [1.40, 1.49]; attribution OR in [0.27, 0.39]), so no single item drives the result. Item fixed effects, which absorb all between-item heterogeneity, leave both effects strong (OR 1.73 and 0.19; both $p<10^{-20}$), and item random intercepts give the same conclusion (1.72 and 0.20). Figure~\ref{fig:item} shows the rise is broad-based across items. The pooled estimates also survive a 1{,}000-replication item-cluster bootstrap (trace-dose 95\% CI [1.30, 1.69]; attribution [0.16, 0.62]). \emph{Verbosity.} Because trace levels also grow in length, two patterns show the effect is not merely verbosity: attribution moves false alarms by 20 to 32 points at fixed trace length, and the longest trace is not the most punitive (for \texttt{qwen3-next-80b}, 98\% at detailed vs.\ 96\% at flawed-verification). The effect tracks asserted verification rather than length. \emph{Decoding.} \texttt{opus-4-7} ran at its provider default rather than at 0.7, so we check that its flatness is not a decoding artifact. It is not decoding deterministically: it returns a different decision across its three runs in 0.99\% of cells, with mean within-cell confidence SD 0.64. Both match \texttt{opus-4-6}, held at 0.7 and also flat (1.00\% and 0.48), and both are far steadier than \texttt{minimax-m2.5} (16.8\%). The two flat overseers behave alike under sampling, so the flat result does not rest on the model whose temperature we did not set.

\section{Discussion}
Three results carry the argument: detection holds while the criterion slides, the slide splits into a removable attribution gap and a residual over-skepticism, and the residual arrives with confidence that does not fall. We take them first for theory, then for design.

\subsection{Theoretical implications}
Our central move is to treat an LLM overseer as a detector and separate its sensitivity from its criterion. Where the evidence is visible, the feared failure, a slick trace fooling the overseer into missing errors, does not occur. What we observe is a \emph{criterion} shift, and because errors are already caught, the extra rejections fall on correct work. A pure accuracy score would misread this as a worse judge, when the judge is no less able, only more suspicious. The two cases have opposite remedies: a sensitivity deficit calls for a more capable model, a criterion problem for recalibrating how much the overseer demands before it rejects.

The decomposition sharpens the contribution. About 60\% of stated rejection reasons are an interface problem, unattributed evidence the overseer cannot tie to a candidate, eliminated to zero by one label. That half is real but contingent: any pipeline that carries a snippet's subject forward is immune to it. The residual does not depend on such a weakness. Under full attribution, a detailed ``I verified everything'' trace still leads susceptible overseers to demand affirmative proof of every criterion, and that is the finding we expect to travel. It connects the faithfulness literature \cite{turpin2023language} to oversight behavior: the \emph{form} of a trace, independent of whether its claims can be checked, moves an AI auditor's decision threshold. It also extends the explanation-as-double-edged-sword finding from human studies \cite{bansal2021whole, vasconcelos2023explanations} to AI overseers, where the edge cuts toward false alarms rather than missed errors, though with no human arm on these items we read that parallel as conceptual. For information systems, the broader implication is that audit trails are decision-shaping governance artifacts. They can recalibrate downstream evaluators and, through them, organizational control outcomes.

\subsection{Managerial and design implications}
In production, false alarms carry a real cost: an over-skeptical overseer escalates valid outputs or sends them back for rework, consuming scarce review capacity and eroding supervisors' trust \cite{lee2004trust, parasuraman2010complacency}. The bite is that the loop exists to relieve human reviewers. An overseer whose criterion has slid returns correct work to the very people it was meant to unburden, recreating the bottleneck the automation was bought to remove. Four design levers follow.

First, label your evidence with what it is about. Tying each retrieved item to the option it describes is a near-free intervention that cut false alarms by up to 32 points. Second, treat explainability as a design variable, not a free good. Piping a verbose ``rigorous'' trace into an overseer can degrade the loop by manufacturing false alarms, and attribution helps but does not fully prevent this; where the evidence pool is available and errors are detectable, concise attributed traces are safer than elaborate ones. Third, separate roles. Two options follow from the mechanism, for overseers that prove susceptible: strip or summarize the worker's trace before it reaches the overseer, and assign trace verification and output judgment to different models, so one model's narrative does not set another's threshold. Fourth, audit the criterion, not just accuracy. Governance should track the false-alarm rate on known-good work and the criterion $c$, not one number that hides whether an overseer is calibrated or merely trigger-happy. The operating point is a cost decision, expected cost $C_{\text{miss}}\,(\text{miss rate}) + C_{\text{FA}}\,(\text{false-alarm rate})$, and our pooled unattributed rates make it concrete. Per 1{,}000 audited items split evenly between correct and flawed work, moving from no trace to a detailed trace takes the miss rate from 1.4\% to 0\% and false alarms from 39\% to 60\%. At equal costs that raises expected cost from 202 to 300 units, and the detailed trace breaks even only once a missed error costs about fifteen times a false alarm. Because over-rejection concentrates in specific overseers, choosing one is itself a control decision.

\subsection{Limitations and future work}
This paper has four limitations. \emph{Scope.} The evidence is 19 compliance items in one domain, with single-criterion violations on structured attributes, and the item is the unit of inference, which is why every estimate is clustered on item and re-checked item by item (Section~5.7). \emph{Detection.} With the disconfirming review always visible, the hit rate sits at ceiling, so the no-gullibility result leaves open whether a trace can mask a genuinely concealed error; a condition with buried errors is the natural test. \emph{Panel.} Five overseers cannot separate scale, vendor, weight availability, and post-training, and the two flat models share a developer, so H4 describes these systems rather than explaining robustness. A larger, more balanced panel including an OpenAI overseer would let susceptibility be modeled rather than observed. \emph{Mechanism.} The decomposition rests on stated reasons, validated against two blind human annotators but still self-reports, and both LLM judges come from one model family. No human overseer was run on these items, so the parallel to the human overreliance literature is conceptual \cite{chen2026synthetic}. Finally, traces are not always harmful. Where evidence is incomplete or errors are subtle a trace may help, and testing criterion-calibration prompts against the residual is the intervention we would try next.

\section{Conclusion}
As organizations increasingly let AI check AI, the worry has been that explanations make overseers gullible. Across 4{,}551 analyzed judgments on compliance tasks where the disconfirming evidence is always visible, we find the opposite. Elaborate procedural traces do not make LLM overseers miss errors; they make susceptible overseers reject correct work, confidently. The over-rejection decomposes into an attribution gap that one label removes and a residual, rigor-driven over-skepticism that survives it, and accuracy-only evaluation cannot see it because it lives in the criterion rather than in sensitivity. Trustworthy AI oversight therefore requires designing for better-calibrated judgment, not for more visible process alone.

\bibliographystyle{unsrtnat}
\bibliography{references}

\end{document}